\documentclass[lettersize,journal]{IEEEtran}
\usepackage{amsmath,amsfonts}
\usepackage{algorithmic}
\usepackage{algorithm}
\usepackage{array}
\usepackage{authblk}
\usepackage[caption=false,font=normalsize,labelfont=sf,textfont=sf]{subfig}
\usepackage{textcomp}
\usepackage{stfloats}
\usepackage{url}
\usepackage{verbatim}
\usepackage{graphicx}
\usepackage{multirow}
\usepackage[
backend=biber,
style=numeric,
sorting=ynt
]{biblatex}

\title{Experimental Insights Towards Explainable and Interpretable Pedestrian Crossing Prediction}

\author[1]{A. N. Melo}
\author[1]{C. Salinas}
\author[1]{M. A. Sotelo}
\affil[1]{Department of Computer Engineering, Universidad de Alcalá, Madrid, Spain \\
[nataly.melo, carlota.salinasmaldo, miguel.sotelo]@uah.es}
\begin{document}

% The paper headers
\maketitle

\begin{abstract}

In the context of autonomous driving, pedestrian crossing prediction is a key component for improving road safety. Presently, the focus of these predictions extends beyond achieving trustworthy results; it is shifting towards the explainability and interpretability of these predictions. This research introduces a novel neuro-symbolic approach that combines deep learning and fuzzy logic for an explainable and interpretable pedestrian crossing prediction. We have developed an explainable predictor (ExPedCross), which utilizes a set of explainable features and employs a fuzzy inference system to predict whether the pedestrian will cross or not. Our approach was evaluated on both the PIE and JAAD datasets. The results offer experimental insights into achieving explainability and interpretability in the pedestrian crossing prediction task. Furthermore, the testing results yield a set of guidelines and recommendations regarding the process of dataset selection, feature selection, and explainability.
\end{abstract}


\section{Introduction}

Nowadays, predicting the behavior of road users holds immense impact, particularly for autonomous driving and intelligent driving systems. These technologies are primarily focused on reducing accidents and enhancing overall road safety. It's crucial to emphasize that pedestrians are among the most vulnerable road users (VRUs), and they constitute the group most significantly impacted on European Union roads. In fact, according to the European Road Safety Observatory and its Annual Accident Report from 2022, a impressive 20\% of fatal accidents involve pedestrians. This alarming statistic has remained constant over the past several years, signifying that one out of every five fatalities on European Union roads involves pedestrians \cite{euSafety}. 

This data underscores the critical need for advancements in pedestrian crossing action prediction and road safety measures to protect this vulnerable group and reduce accidents on the road. In light of this, many research communities have been developed Machine Learning (ML) models and methods to make more robust prediction systems and to face the related challenges.

For a long time most of these models and methods were viewed as \emph{'black boxes'} because they lacked the ability to explain the reasoning behind their predictions. Consequently, understanding why a machine learning model made a specific prediction was challenging. Besides, it is widely recognized that, especially for certain tasks is not enough to obtain the prediction, and equally crucial is the capability to interpret that prediction. 

In the context of autonomous driving, this challenge has gained prominence due to the imperative need to clarify every decision made by autonomous vehicles (AVs). In \cite{avexplainable}, the author highlights the necessity for explanations from multiple perspectives:
\begin{itemize}
    \item Psychological Perspective: Understanding why a specific decision was made can have deep implications for enhancing road safety and reducing traffic accidents.
    \item Sociotechnical Perspective: This viewpoint recognizes that the design and development of autonomous driving systems must be human centred.
    \item Philosophical Perspective: Explaining ego-vehicle decisions contributes to providing descriptive information regarding the causal history of actions taken
\end{itemize}

In this work, we propose an experimental neuro-symbolic approach to develop an explainable and interpretable pedestrian crossing predictor. We base this predictor on fuzzy logic and deep learning for feature extraction. The proposed method employs multiple explainable features extracted from the JAAD and PIE datasets for mining fuzzy rules. Subsequently, these rules are used to define a fuzzy inference system that facilitates pedestrian crossing prediction. This approach is particularly novel as it not only focuses on accurate prediction but also establishes a baseline for explainability.

On the other hand, it's crucial to acknowledge that the success of machine learning projects is significantly influenced by the quality and relevance of the datasets used for model training and testing. In fact, a well-chosen dataset has the potential to enhance the accuracy and efficiency of the model, whereas an inaccurate selection can yield unfavorable results. Furthermore, the characteristics of the datasets play a crucial role in shaping the behavior of a model. It is essential to consider that a model's performance in real-world scenarios may be compromised if its deployment context significantly differs from the training and evaluation datasets \cite{datasheet}.
Furthermore, it is important to highlight that within the context of explainability and interpretability, the selection of the dataset holds significant relevance. This is because a well-chosen dataset can play a key role in facilitating the extraction of explainable features from pedestrian crossing action datasets.

The rest of the paper proceeds as follows: Section 2 discusses the related work. Details about the proposed architecture for an explainable pedestrian action predictor, feature selection and fuzzy logic definitions are introduced in Section 3. Section 4 describes the implementation and experimental setup. The experimental results and analysis are presented in Section 5. Section 6 provides a baseline for explainability of action pedestrian predictors. The guidelines and recommendations regarding feature selection, dataset selection and explainable prediction are introduced in Section 7. Finally, in Section 8 are presented the conclusions and future work. 

\section{Related Work}
\subsection{Approaches for dataset selection}
Datasets are playing an important role into the ML Projects, with researchers actively exploring significant aspects and considerations related to them. For instance, there are studies that aim to comprehensively document the creation and utilization of datasets through the development of datasheets. These datasheets provide valuable information regarding the motivation behind dataset creation, composition details, collection processes, preprocessing techniques, distribution methods, besides data usage and maintenance guidelines \cite{datasheet}. Similarly, research efforts are being dedicated to various aspects of datasets, such as tracking and controlling dataset versions \cite{versions} and exploring data provenance \cite{provenance}. 

In addition, there are some data selection methods which focus on choosing the most informative training examples for machine learning tasks across a specific dataset \cite{dataselection}. Nevertheless, it is worth noting that the topic of \textbf{dataset selection} approaches is relatively under-discussed within ML research community. 

\subsection{Pedestrian Action Datasets}
In the context of autonomous driving, numerous datasets encompass pedestrian annotations. While datasets like Trajectory Inference using Targeted Action priors Network (TITAN) \cite{titan} concentrate on a range of pedestrian actions, including motion, communicative, and contextual actions, they do not explicitly incorporate the task of crossing action in their dataset approach. Conversely, datasets such as Stanford-TRI Intent Prediction (STIP)\cite{stip}, the Joint Attention for Autonomous Driving (JAAD) \cite{Jaad}, and Pedestrian Intention Estimation (PIE) \cite{pie} explicitly include crossing actions within their dataset approach.

Taking this into consideration, for our initial exploration into an explainable and interpretable pedestrian action predictor, we have chosen to focus on widely used and publicly available datasets within the research community, namely JAAD and PIE.
\subsubsection{JAAD Dataset}
JAAD is a richly annotated datataset composed by 348 short video clips. It includes a diverse number of road actors for each scene and several driving locations, traffic and weather conditions. The dataset's annotations are divided by spatial, behavioural, contextual and pedestrians information. \\
Regarding the pedestrian action annotated in JAAD, it is noteworthy that approximately 72\% of pedestrians actively cross the street, whereas the remaining 28\% do not engage in street crossing. 

\subsubsection{PIE Dataset}
The creators of JAAD released a new dataset called PIE, it contains over 300K labeled video frames recorded in Toronto in clear weather. 
In addition to the similar type of annotations found in JAAD, PIE stands out by incorporating ego-vehicle information derived from on-board diagnostics (OBD) sensors. 
Distinguished from the JAAD dataset which primarily focuses on pedestrians intending to cross, the PIE dataset provides annotations for all pedestrians in close proximity to the road, irrespective of whether they attempt to cross in front of the ego-vehicle or not. Furthermore, in contrast to the annotated pedestrian action in the JAAD dataset, the PIE dataset exhibits as well an imbalance where a larger proportion of pedestrians are observed not crossing the street. Specifically, approximately 39.2\% of pedestrians actively cross the street, whereas the remaining 60.8\% do not engage in street crossing. 

\subsection{Explainability and interpretability concepts}
In general, there is not a consolidated agreement within the ML community on the definition of interpretability and explainability and many authors use these terms interchangeably  \cite{explainability}. However, some authors as Lipton in \cite{questionsExp}, emphasize the distinction between these concepts by framing them as questions: interpretability raises the question \emph{"How does the model work?"} while explainability attempts to answer \emph{"What additional insights can the model provide?"} \cite{explainability}. Additionally, the term interpretability is often associated with the degree to which a human can comprehend the rationale behind a decision, while explanation refers to the response to a why question \cite{interbook}. In this paper we will use both terms as complementary for each other. 

\subsection{Pedestrian Crossing Action Prediction and its explainability}
Pedestrian crossing action prediction is a ML task focused on forecasting if a pedestrian will cross the road at some point in the future. This task has been addressed through a diverse range of algorithms and architectures. Among these approaches, it is particularly noteworthy to highlight the SingleRNN method that focuses on leveraging contextual features and employing an encoder-decoder architecture powered by recurrent neural networks (RNNs) \cite{crossrnn}. In \cite{gnpose}, the authors used a graph-based model and 2D human pose estimation to predict whether a pedestrian is going to cross the street. The method of CapFormer \cite{capformer} uses a self-attention alternative based on transformer architecture. This method primarily focuses on bounding boxes, pose estimation, and AVs speed. Additionally, a 3D convolutional model (C3D) is employed, along with cropped regions of pedestrian bounding boxes from RGB video sequences, to facilitate spatiotemporal feature learning \cite{c3d}. Another group of algorithms relies on stacked with multilevel fusion RNN (SFRNN) \cite{stackrnn} and convolutional LSTM (ConvLSTM) \cite{convlstm}. 

Furthermore, it is worth mentioning the recently developed benchmark by York University, which extensively assessed the performance of various ML approaches in the pedestrian crossing task. This benchmark not only standardized the evaluation criteria for the task but also introduced a model that combines the power of RNNs and 3D convolutions \cite{benchcrossing}. 

On the other hand, despite the abundance of models and research focused on pedestrian crossing predictions, only a limited number of them provides insights into explainability or are specifically developed within the context of explainability. For instance, the research \cite{transformerexplain} highlights that Transformers offer an advantage in terms of interpretability, due to their attention mechanism. Moreover, the utilization of pedestrian location and body keypoints as features in predicting pedestrian actions results in more human-like behavior. In \cite{emidas}, the authors propose a dynamic Bayesian network model that takes into account the influence of interaction and social signals. This system leverages visual means and employs various inference methods to provide explanations for its predictions, with a specific focus on determining the relative importance of each feature in influencing the probability of pedestrian actions.

\subsection{Fuzzy Logic}
Fuzzy logic was introduced by Zadeh \cite{fuzzyset} in order to deal with "degrees of truth" rather than absolute values of "0 and 1". It can be defined as a multi-valued logic that closely resembles human thinking and interpretation. The main components of fuzzy logic are \cite{fuzzyneuro}\cite{fuzzyset}:
\begin{itemize}
    \item Fuzzy sets. Composed by linguistic variables where values are words and not numerical.
    \item Membership functions. Define the shape and characteristics of fuzzy sets.
    \item Fuzzy rules. Connects various input and output fuzzy variables through conditional statements expressed as "if...then" rules.
    \item Fuzzy reasoning. Draws the conclusions from fuzzy sets and fuzzy rules
\end{itemize}

Fuzzy inference system (FIS) is a framework based on fuzzy logic components that follows a step-by-step procedure to provide outputs that can be explained. The mentioned procedure implies different steps such as fuzzifier, rule base, inference engine and defuzzier \cite{fuzzyneuro}. These steps collectively form the FIS framework, which enables the processing of fuzzy inputs to produce understandable and interpretable outputs. 
It is important to highlight that fuzzy logic  is a solution to complex problems which requires human reasoning and decision making. It has been widely used in the field of Intelligent Transportation Systems (ITS), specially in automated vehicle control \cite{fuzzyits} \cite{fuzzyits2}. In addition, its application extends to various areas as mentioned in \cite{surveyfuzzy}, including driver behavior modeling, analyzing alternate routes, predicting traffic patterns, and addressing traffic control issues.

\section{Explainable Predictor Proposal}
The following section introduces the general approach proposed to achieve an explainable pedestrian crossing predictor based on fuzzy logic. In addition the selected features are briefly explained.

\subsection{Proposed Approach}
\begin{figure}[h!]
  \centering
  \includegraphics[scale=0.50]{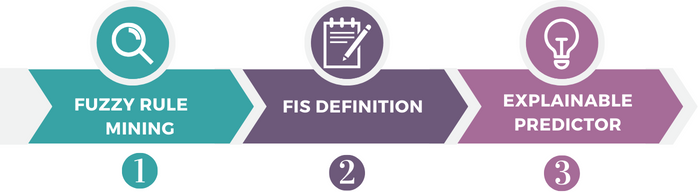}
  \centering
  \caption{Experimental approach steps}
  \label{fig:approach_steps}
\end{figure}

In this work, we proposed an experimental approach to develop an explainable and interpretable pedestrian crossing predictor (ExPedCross) based on a neuro-symbolic model using fuzzy logic. The proposed approach is divided into three main steps: (1) fuzzy rule mining, (2) fuzzy inference system definition and (3) explainable predictor (see Figure \ref{fig:approach_steps}).

\begin{figure}[h!]
  \centering
  \includegraphics[scale=0.5]{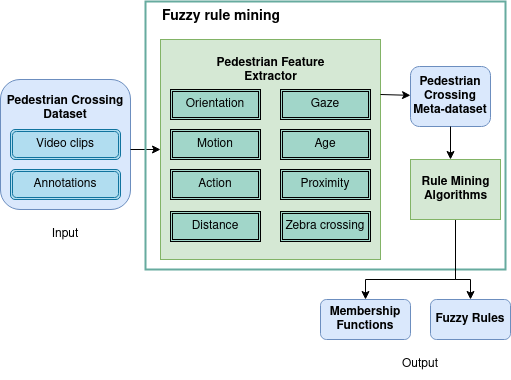}
  \centering
  \caption{Step 1: Fuzzy rule mining}
  \label{fig:fuzzy_mining}
\end{figure}

The rule mining process (Step 1) involved the potential of feature extraction from on side, and the utilization of diverse algorithms to generate fuzzy rules on the other side. As can be described in the Figure \ref{fig:fuzzy_mining}, the rule mining process begins with pedestrian action dataset (JAAD and PIE) consisting of videos, images and annotations, then a feature extractor component is employed to extract features using deep learning and neural networks, creating a meta-dataset that contains the extracted features for each pedestrian (The description of these features is detailed in section III). Then the meta-dataset is used to extract fuzzy rules and membership functions through fuzzy rule learning algorithms. It is important to highlight that the data used for extracting fuzzy rules maintains a balance between the number of pedestrians who cross the street and those who do not. This balanced representation ensures that the extracted fuzzy rules are equally informed by both scenarios.

For the rule mining process several fuzzy rule learning algorithms were evaluated , including notable ones such as  Chi-RW-C \cite{chirw}, GP-COACH-C \cite{gpcoach} and IVTURS-FARC \cite{ivturs}. Among them, IVTURS-FARC was selected based on its performance and promising preliminary results in generating fuzzy rules from the meta-dataset. IVTURS-FARC is a linguistic fuzzy rule-based classification algorithm which uses interval-valued restricted equivalence functions to increase the relevance of the rules during the inference process. The fuzzy rule learning process of IVTURS-FARC utilizes the FARC-HD algorithm \cite{farchd}.  The resulting fuzzy rules take the following form, as described in the study \cite{ivturs}:
\begin{equation}
    \begin{split}
        Rule \; R_{j} : & If \;x_{1}\; is \;A_{j1}\; and ... and\;  x_{n}\; is\; A_{jn} \\
                     & then\; Class\; = C_{j}\; with\; RW_{j}
    \end{split}
\end{equation}
where \(R_{j}\) is the label of the \(j\)th rule, \(x = (x_{1},...,x_{n})\) is an n-dimensional pattern vector (pedestrian features in our work), \(A_{ji}\) is an antecedent fuzzy set representing a linguistic term, \(C_{j}\) is the class label and \(RW_{j}\) is the rule weight \cite{rulesdef}.  
\\
The second step of the proposed approach uses the output of rule mining process to define the fuzzy inference system. This system has been defined as Takagi-Sugeno (TS) fuzzy inference \cite{TS} which allows to represent non-linear systems with a set of fuzzy rules of which consequent parts are linear state equations. In our proposal the TS fuzzy inference system included as an input the fuzzifies pedestrian features with its membership function, the generated rules and zero-order output values. To have a clear picture of the fuzzy inference procedure, we defined the crossing prediction output \(C_{j}\) according to :

\begin{equation}
    C_{j}= \left\{ \begin{array}{lr} Cross & : 1\\ Not Cross & :  0 \end{array} \right.
\end{equation}
In the last step (Step 3) , we have the explainable pedestrian crossing predictor (ExPedCross) that uses fuzzy logic and a neuro-reasoning approach to predict whether a pedestrian will cross the street. By incorporating pedestrian features as input and employing the TS fuzzy inference system, this novel approach offers a high level of explainability through the activation of fuzzy rules during each prediction. For the initial experiments, the ExPedCross uses 1 frame as observation time to predict the next 30th frame.

\subsection{Features Selection}
In this study, we carefully choose eigth features to serve as inputs for ExPedCross. Among these features, seven are derived through the implementation of neural networks, while one feature is obtained directly from the pedestrian crossing datasets. The following list outlines the selected features:
\begin{itemize}
    \item \textbf{Motion Ability} \((MA_{P}) \) describes the motion capability of the pedestrians: Fully capable, using wheelchair, using crutches, using walking frame and pushing a wheelchair.
    \item \textbf{Age} \((AG_{P}) \) describes the pedestrian age considering two classes: Adult and child.
    \item \textbf{Body Orientation} \((BO_{P}) \) describes the pedestrian posture through an angle from 0 to 360º.
    \item \textbf{Gaze} \((GA_{P}) \) describes the attention of the pedestrian, indicating whether the pedestrian is looking at the ego-vehicle.
    \item \textbf{Action} \((AC_{P}) \) describes the motion state of the pedestrian, classifying between the following actions: stand, walk, wave, run or undefined (used when pedestrian action is not clear).
    \item \textbf{Proximity to the road} \((PR_{P}) \) describes if the pedestrian is near to the road. This feature is classified in three levels according the pedestrian closeness to the road: near, medium distance or far.
    \item \textbf{Zebra Crossing} \((ZC_{P}) \) represents the presence of a zebra crossing in the scene.
    \item \textbf{Distance} \((DE_{P}) \) represents the estimated distance to the ego-vehicle.
\end{itemize}

Regarding the experiments conducted in this work, it is important to highlight that in these initial experiments, the features used specifically focus on capturing the pedestrian state on a frame-by-frame basis. This means that the features are extracted and analyzed for each individual frame, providing a detailed understanding of the pedestrian's state at each moment in time. 
Furthermore, in this paper, the features 'motion ability' and 'age' will not be considered. According to the experimental results, these features did not significantly contribute to the performance of ExPedCross. This can be attributed to the lack of sufficient samples representing a diverse range of pedestrian motion capabilities and ages in the JAAD and PIE datasets. 

\section{Implementation and Experimental Setup}
In this section, we detail the implementation and the test methodology to prove the ExPedCross.

\subsection{Data Sampling}
In this work, we split the datasets mentioned previously for two main tasks: (1) training during the fuzzy rule mining step (See Figure \ref{fig:fuzzy_mining}) and (2) testing the performance of  ExPedCross.

Besides to ensure the reliability of our experiments, we initially selected videos that met specific criteria such as visibility and high quality. As a result, we carefully choose 284 videos from JAAD and 53 videos from PIE to be included in this experiment. 

In addition during the training phase, the videos were carefully sorted based on their quality and relevance, prioritizing the most representative ones at the beginning of the training set. This sorting strategy aimed to enhance the learning process by focusing on the most informative videos.

For the testing phase, we defined four distinct groups: (1) \(JAAD_{all} \), (2) \(JAAD_{beh} \),  (3) \(PIE_{all} \) and (4) \(PIE_{beh} \). The \(_{all} \) versions include all the selected videos from the JAAD and PIE dataset, while the \(_{beh} \) group excludes videos with pedestrians annotated as irrelevant during the meta-dataset generation. 

By splitting the datasets and defining these groups, we ensured a comprehensive evaluation of our model's performance across different scenarios, including both relevant and irrelevant pedestrian annotations. Afterwards, the data sampling mentioned above was used to generate the meta-dataset from JAAD and PIE datasets. In addition, it is important to highlight that due to the imbalance between pedestrians who cross the street and those who do not, special attention was given to creating a balanced meta-dataset. Besides, two rules were defined who support carefully the selection of data, trying to reduce data noise: (1) Don't consider more than 60 frames after pedestrian cross, (2) Don't consider more than 90 frames when the pedestrian will not cross. 

\subsection{Features Extractor}
The features extractor was developed with a modular architecture which allows to include different modules to extract each feature. It was developed using python and PyTorch; besides it is composed by 8 modules which are responsible to extract each feature as is described:
\\
\subsubsection{Pedestrian Motion Ability and Age}
To extract the pedestrian ability feature we used the object detection approach, focusing on a You Only Look Once (YOLO), specifically we use YOLOv7 \cite{yolov7}. To train this convolutional neural network (CNN) we decide to create a custom dataset from sourced images from Google Images, Pexels and freePick. This dataset is composed by 1498 images distributed in 7 classes such as adult, child, wheelchair, crutches, walking frame and wheelchair pusher. Hence, this module receives an image as input and provides three key outputs: the motion ability, pedestrian age, and the corresponding bounding box.
\subsubsection{Pedestrian Orientation}
To calculate the pedestrian body orientation it was used the PedRecNet \cite{pedrecnet} network designed and implemented by the Reutlingen University. This neural network is a multitask network that supports various pedestrian detection functions from 2D and 3D human pose. Based on the joint positions it is estimated the human body orientation, generating as a result the polar angle \( \theta \)  and the azimuthal angle \( \varphi \). Human body orientations are output as angles between 0º - 180º for \( \theta \)    and 0º - 360º for \( \varphi \).

The \( \varphi \) angle allows to describe if the pedestrian is oriented to the left, the Right, is in the car direction or opposite to car direction. Nevertheless, the angles which can indicate Right direction are in a discontinuous range 0º a 45º y 315º a 0º.
Taking into account that a fuzzy set is characterized by a membership function \(f_{A}(x) \) represented by continuum truth values in an interval and considering that the mentioned discontinuous range can not be represent as a fuzzy set, we decide to shifted +45º the final orientation.
\subsubsection{Pedestrian Gaze}
To calculate if the pedestrian is looking at the vehicle, it is used the 2D body pose detected through the PedRecNet \cite{pedrecnet}. Pedestrian gaze is determined by focusing on position of the nose, left eye, and right eye keypoints \cite{brave}. 
\subsubsection{Pedestrian Action}
To detect the pedestrian action it was implemented a transformer inspired by the Action Transformer \cite{actiontransformer}. The transformer architecture implemented focuses only in the encoder part. The input of the neural network is an feature array with shape B x F x  K , where B represent the batch size, F represents the frames and K contains the pedestrian's body pose reflected as keypoints. The output is a number which represents the pedestrian action.

To train the transformer it was used the MPOSE2021 dataset \cite{actiontransformer} which relates the human skeleton with human action. This dataset is composed of 20 distinct actions. However, in order to focus on the actions most relevant for pedestrian crossing predictions, we made the decision to reduce the number of classes. Upon comparing the test results, we observed that using 20 classes during training resulted in a 90\% accuracy rate on the testing set. Surprisingly, when we reduced the classes to just 5, the accuracy improved significantly to 94\%. Therefore, we proceeded to condense the dataset into these 5 classes, which are grouped as illustrated in the following table:
\begin{table}[h!]
    \centering
    \caption{Pedestrian actions group}
    \label{table:1}
    \begin{tabular}{ | m{2cm} | m{5cm} | }
        \hline
         Class & Values  \\ 
        \hline
         0: Standing & Standing, Check-watch, Cross-arms, Scratch-head, Hands-clap  \\ 
        \hline
         1: Walk & Walk, Turn \\ 
         \hline
         2: Wave & Wave, Point, Wave2 \\ 
         \hline
         3: Run & Run, Jog \\ 
         \hline
         4: Undefined & Sit-down, Get-up, Box, Kick, Pick-up, Bend, Jump, Position Jump \\ 
         \hline
    \end{tabular}
    
\end{table}

\subsubsection{Pedestrian Proximity to the road}
To detect the pedestrian proximity to the road it was used a YOLOPv2 pretrained network \cite{yolopv2}. YOLOPV2 is a multi-task learning network that performs the task of traffic object detection, drivable road area segmentation and lane detection. Based on the drivable road area segmentation and lane detection  and an experimental minimum distance it is estimated if the pedestrian is near to the road or not.

\subsubsection{Pedestrian Distance}
The pedestrian distance was estimated using the triangle similarity, which is represented in the equation 3, where W is a known width of the pedestrians, F is the focal length and P is the pedestrian width in pixels:
\begin{equation}
 D = (W x F ) /P
\end{equation}
It is important to highlight that for each dataset it was required to compute a different estimation, due to the difference between the camera parameters used for each record.

\subsection{Explainable Pedestrian Crossing Predictor}
The development of the ExPedCross predictor involved two main phases. In the first phase, we utilized the KEEL software \cite{keel}, an open-source Java framework, to generate the fuzzy rules from a pedestrian meta-dataset. 

\begin{figure}[h!]
  \centering
  \includegraphics[scale=0.3]{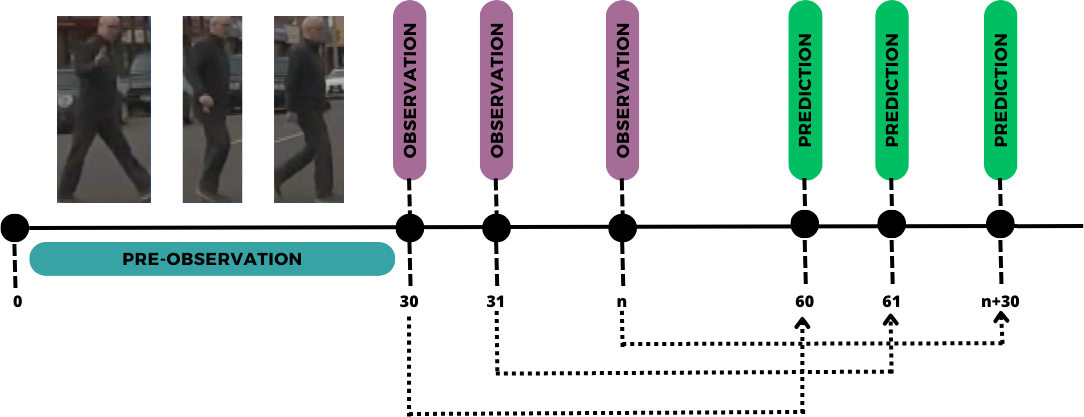}
  \centering
  \caption{ExPedCross timing}
  \label{fig:timing}
\end{figure}

In the subsequent phase, we developed a software application that takes as input a vector with a shape of 1 x 8, encompassing the pedestrian features extracted from 1 after the first 30 frames of observation (See Figure \ref{fig:timing}). Based on this vector the crossing behaviour of the next \(30_{th}\) was predicted using fuzzy logic. This software was developed in python and using the simpful \cite{simpful} library.

\subsection{Test Methodology}
To understand the significance of dataset and data selection, we conducted a series of experimental tests on our explainable predictor. In the context of ML, factors such as dataset quantity, data preprocessing, randomness, ablation factor, and the incorporation of diverse datasets play important roles. Therefore, each factor was defined as an experiment. The paper evaluate the result of 5 experiments.

The experiments are carried out by using different configurations, each addressing specific aspects of dataset and data selection. The order of experiment's execution was determined with the intention of obtaining valuable insights and feedback for subsequent experiments. Regarding the evaluation of the experiments we use the f1 score (F1) metric. In summary, F1 Score is the harmonic mean between precision and recall. 

In light of this, the test methodology follows the next sequence of instructions:
\begin{itemize}
    \item Sort the video list for JAAD and PIE taking into account the quality and usability of each video.
    \item For each experiment:
    \begin{itemize}
    \item Define the configurations to test in JAAD and PIE
    \item For each configuration:
    \begin{itemize}
        \item Use the fuzzy rule mining process (See Figure \ref{fig:fuzzy_mining}) to generate the fuzzy rules and membership functions. The input data of this process should vary taking into account the test configuration.
        \item Use the fuzzy rules generated and the ExPedCross to predict and mesure the results over the four groups of testing: (1) \(JAAD_{all} \), (2) \(JAAD_{beh} \),  (3) \(PIE_{all} \) and (4) \(PIE_{beh} \).
    \end{itemize}
    \item Analyze the test results and provide the feedback
    \end{itemize}
\end{itemize}

\section{Experiments Over An Explainable Predictor}
The following section presents the experimental results focused on pedestrian crossing prediction task.
\subsection{Quantity Factor}
In the context of ML, one of the critical decisions when training a model is determining the amount of data to use. This decision can vary depending on the specific ML task at hand. The initial experiments were conducted to address the following question: Does having more data result in better performance of the predictions? 

To address this question, we designed a set of configurations that incrementally increased the number of data records for PIE and JAAD. We initially started with 2000 records, balanced equally between pedestrians who cross and don't cross. Subsequently, we expanded the dataset to include 8000 rows, 10,000 rows and then 14,000 rows, gradually increasing the volume of data. Since the PIE dataset is larger in size, we extended the analysis to incorporate two additional configurations. These configurations included 20,000 rows and 80,000 rows, specifically for the PIE dataset. 

The following table presents the result of the different configurations mentioned above:
\begin{table}[h!]
    \centering
    \caption{Results Quantity Factor. Abbr. Conf:Configuration, R:Rules, F1: F1 score, J:JAAD dataset , P:PIE dataset}
    \label{table:quantity}
\begin{tabular}{lllllr}
\hline & & \(JAAD_{all} \) & \(JAAD_{beh} \) &
\(PIE_{all} \) & \(PIE_{beh} \)\\
\cline{3-6}
Conf    & R & F1 & F1 & F1 & F1  \\
\hline
J2K  & 55  & 0.70  & 0.71  & 0.44  & 0.48   \\
J4K  & 68  & 0.72  & 0.72  & 0.53  & 0.56   \\
J8K  & 62  & 0.74  & 0.75  & 0.50  & 0.53   \\
J10K & 73  & 0.74  & 0.75  & 0.46  & 0.50   \\
J14K & 58  & \textbf{0.75}  & \textbf{0.76}  & 0.46  & 0.51   \\
P2K  & 38  & 0.46  & 0.47  & 0.48  & 0.51   \\
P4K  & 49  & 0.57  & 0.57  & 0.49  & 0.52   \\
P8K & 54  & 0.58  & 0.59  & 0.45  & 0.52   \\
P10K & 45  & 0.56  & 0.55  & 0.46  & 0.52   \\
P14K & 49  & 0.64  & 0.66  & 0.46  & 0.53   \\
P20K & 50  & 0.57  & 0.58  & 0.47  & 0.51   \\
P80K & 36  & 0.59  & 0.60  & \textbf{0.54}  & \textbf{0.57}   \\
\hline
\end{tabular}
\end{table}

By systematically expanding the dataset in this manner, we can identify that more amount of data can improve the performance of the predictor when it is testing over the same dataset. However, it does not imply better results over other datasets. This means that for an explainable predictor based on fuzzy logic, increasing the amount of data does not guarantee better generalization on its own. Furthermore, as can be observed in the Table \ref{table:quantity}, more rules do not necessarily provide better results, because one of the best configuration provides the minimum number of rules over all the configurations defined: 36 rules defined in the P80K configuration.

In addition it is important to mention that the configuration \textbf{J14K} based on JAAD contains in average the best result of the experiments. This configuration contains 66,000 records less than the second best configuration P80K based on PIE. That can be explained because JAAD dataset contains more diverse scenarios and pedestrians which can allow to understand better the behaviour and characteristics of the pedestrian related with the cross/not cross action. 

\subsection{Randomness factor}
The objective of this experiment was to understand if including a human reasoning over the selection videos to get the fuzzy rules, can improve the performance of the explainable predictor or if the randomness selection is enough to get accurate results. 
This experiment takes as baseline the result from the configuration J8K and P14K from the Quantity factor experiment. As was described in the test methodology, prior to selecting the data input for the fuzzy rule mining process, the videos were listed and ordered for each dataset according its quality and usability on the pedestrian crossing prediction context. Then, this experiment includes 6 more configurations which are compose of the same number of records for each case (8,000 and 14,000 records) but where the selection of videos is random.

According to the results present in the Table \ref{table:random}, the random selection of the data could improve the performance of explainable predictor over its own dataset.  Configurations J8KR3 and P14KR1 have better results over JAAD and PIE respectively. However, the process of analyzing the videos and sorting them according to a classification that evaluates relevance, quality, and context of the video, can improve the performance and generalization of the explainable predictor. In fact, configurations J8K and P14K present better results over cross-testing.

\begin{table}[h!]
    \centering
    \caption{Results Randomness Factor. Abbr. Conf:Configuration, R:Rules, F1: F1 score, J:JAAD dataset, P:PIE dataset }
    \label{table:random}
\begin{tabular}{lllllr}
\hline & & \(JAAD_{all} \) & \(JAAD_{beh} \) &
\(PIE_{all} \) & \(PIE_{beh} \)\\
\cline{3-6}
Conf    & R & F1 & F1 & F1 & F1  \\
\hline
J8K    & 62  & 0.74  & 0.75  & \textbf{0.50}  & \textbf{0.53}   \\
J8KR1  & 66  & 0.68  & 0.70  & 0.34  & 0.40   \\
J8KR2  & 80  & 0.75  & 0.75  & 0.46  & 0.50   \\
J8KR3  & 70  & \textbf{0.76}  & \textbf{0,77}  & 0.48  & \textbf{0.53}   \\
\cline{1-6}
P14K   & 49  & \textbf{0.64}   & \textbf{0.66}  & 0.46  & 0.53   \\
P14KR1 & 59  & 0.62  & 0.62  & \textbf{0.51}   & \textbf{0.55}    \\
P14KR2 & 28  & 0.39  & 0.39  & 0.50  & 0.52   \\
P14KR3 & 57  & 0.57  & 0.57  & 0.48  & 0.51   \\
\hline
\end{tabular}
\end{table}

\subsection{Selection Factor}
This experiment was focused on understanding the impact and importance of selecting adequate data to obtain accurate results from an explainable predictor. The main objective was to confirm the hypothesis that careful data selection is crucial for obtaining meaningful fuzzy rules. Therefore, the experiment explored different and incremental configurations generated without any frame restrictions (See Section IV-A). In fact, the configuration for this experiment includes the following configurations: For JAAD: (1) The J14K from Quantity experiments, (2) J14Kn with 14,000 records without data filtered and (3) J30Kn with 30,000 records without data filtered. From PIE: (1) P80K from Quantity experiments, (2) P80Kn with 80,000 records without data filtered and (3) J160K with 160,000 records without data filtered.
\begin{table}[h!]
    \centering
    \caption{Results Selection Factor. Abbr. Conf:Configuration, R:Rules, F1: F1 score }
    \label{table:selection}
\begin{tabular}{lllllr}
\hline & & \(JAAD_{all} \) & \(JAAD_{beh} \) &
\(PIE_{all} \) & \(PIE_{beh} \)\\
\cline{3-6}
Conf    & R & F1 & F1 & F1 & F1  \\
\hline
J14K  & 58  & \textbf{0.75}  & \textbf{0.76}  & 0.46  & 0.51  \\
J14Kn  & 28  & 0.72  & 0.73  & 0.42  & 0.46   \\
J30Kn  & 29  & 0.71  & 0.71  & 0.44  & 0.48   \\
\cline{1-6}
P80K & 36  & 0.59  & 0.60  & \textbf{0.54}  & \textbf{0.57}   \\
P80Kn & 79  & 0.60  & 0.61  & 0.50  & 0.54   \\
J160Kn  & 69  & 0.42  & 0.42  & 0.50  & 0.51   \\
\hline
\end{tabular}
\end{table}

According to the results presented in the Table \ref{table:selection},  the selection and filtering of data had a significant impact on fuzzy rule generation. The rules created from data with certain frame restrictions demonstrated better performance for the explainable predictor.

\subsection{Ablation Factor}
This experiment was focused on understanding the relationship between the different  features and the prediction results. It builds upon the best results obtained from the quantity factor experiments, specifically the configurations J14K and P80K. In this experiment, the data input for the fuzzy rule mining process was modified. The changes involved eliminating each feature from configuration J14K and P80K, to assess its impact on the prediction results. 
The following table presents the result of the different configurations mentioned above:
\begin{table}[h!]
    \centering
    \caption{Results Features Factor. Abbr. Conf:Configuration, R:Rules, F1: F1 score, J:JAAD dataset, P:PIE dataset }
    \label{table:features}
\begin{tabular}{lllllr}
\hline & & \(JAAD_{all} \) & \(JAAD_{beh} \) &
\(PIE_{all} \) & \(PIE_{beh} \)\\
\cline{3-6}
Conf    & R & F1 & F1 & F1 & F1  \\
\hline
J14K           & 58  & \textbf{0.75}  & 0.76  & 0.46  & 0.51   \\
J14K-NDistance    & 56  & 0.74  & 0.75  & 0.41  & 0.45   \\
J14K-NProximity   & 45  & 0.49  & 0.50  & 0.30  & 0.34   \\
J14K-NAction      & 50  & \textbf{0.75}  & 0.75  & 0.43  & 0.49   \\
J14K-NAttention   & 54  & \textbf{0.75}  & \textbf{0.77}  & 0.45  & 0.51   \\
J14K-NOrientation & 46  & 0.74  & 0.76  & \textbf{0.48}  & \textbf{0.54}   \\
J14K-NZebraCross       & 51  & 0.74  & 0.75  & 0.47  & 0.52   \\
\cline{1-6}
P80K           & 36  & 0.59  & 0.60  & 0.54  & 0.57   \\
P80K-NDistance    & 36  & 0.58  & 0.58  & 0.43  & 0.44   \\
P80K-NProximity   & 31  & 0.53  & 0.54  & 0.54  & 0.57   \\
P80K-NAction      & 17  & 0.47  & 0.48  & 0.51  & 0.55   \\
P80K-NAttention   & 25  & 0.58  & 0.59  & 0.54  & \textbf{0.59}   \\
P80K-NOrientation & 15  & 0.53  & 0.54  & \textbf{0.55}  & 0.58   \\
P80K-NZebraCross   & 22  & \textbf{0.60}  & \textbf{0.61}  & 0.53  & 0.58   \\
\hline
\end{tabular}
\end{table}

The most representative feature extracted from JAAD is proximity, while from PIE  are distance and action. This conclusion is drawn from observing the reduction in prediction performance when these features are eliminated from the input (J14K-NProximity and JP80K-NDistance configurations) during the fuzzy rule learning process. The significant decrease in prediction accuracy without these features suggests their importance in the overall prediction process.

Furthermore, during cross-testing, it is important to note that the orientation feature from JAAD and the label of zebra crossing from PIE can have a negative impact on the results. This implies that the representation of these features varies considerably between the two datasets. The divergent representation of these features across datasets may lead to inconsistencies in prediction performance when the explainable predictor is applied to different environments or scenarios.

To complement this experiment, we conducted an analysis of the correlation between all the selected features and the pedestrian crossing action. Specifically, we focused on the data from the configurations J14K and P80K for this correlation analysis. 

\begin{figure}[h!]
  \centering
  \includegraphics[scale=0.55]{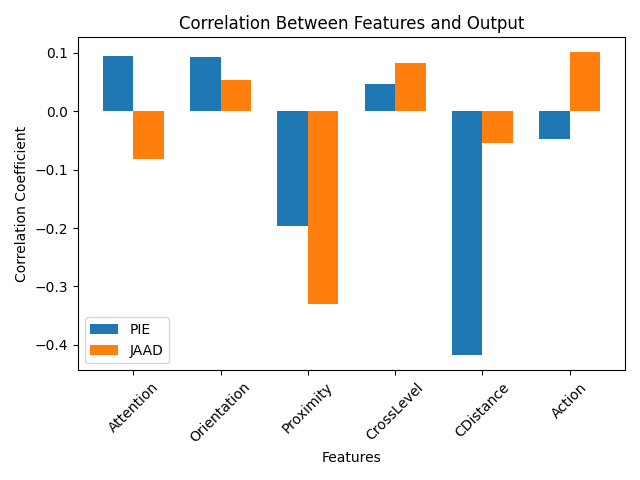}
  \centering
  \caption{Features Correlation}
  \label{fig:features_correlation}
\end{figure}

In Figure \ref{fig:features_correlation}, it is evident that pedestrian crossing action shows a small positive correlation with the pedestrian action in the JAAD dataset. This suggests that as the pedestrian's action class increases (as seen in table \ref{table:1})), there is a higher likelihood of the pedestrian crossing. Additionally, a medium negative correlation is observed between the pedestrian's proximity to the road and the likelihood of crossing in both the JAAD and PIE datasets. This implies that when the pedestrian is in close proximity to the road, there is a greater likelihood of them crossing. Lastly, in the PIE dataset, a medium negative correlation is observed between the crossing action and the pedestrian's distance. This suggests that when the pedestrian's distance to the ego-vehicle is low, there is a greater likelihood of them crossing.

Conversely, the remaining pedestrian features do not exhibit a significant correlation with the pedestrian's decision to cross. These findings suggest that, in general, the evaluated features alone do not provide sufficient representatives for predicting pedestrian crossing behaviour. Instead, they need to be complemented with additional features that convey more representative information.

\subsection{Mix factor}
The main goal of this experiment was to identify if the explainable predictor performance improves by mixing datasets for training. To develop the experiment  three configurations were generated which combined JAAD and PIE as a data input from the fuzzy rule mining process. The configurations were defined as follow: (1) J8K-P8K: 8,000 records from each dataset, (2) J10K-P10K: 10,000 records from each dataset and (3) J14K-P80K 14,000 records from JAAD and 80,000 records from PIE.

According to the result presented in the Table \ref{table:mix} when the datasets are combined to create the data input, there is only a slight improvement in performance during cross-testing. This behaviour can be attributed to the fact that the datasets share similar scenarios.

\begin{table}[h!]
    \centering
    \caption{Results Mix Factor. Abbr. Conf:Configuration, R:Rules, F1: F1 score}
    \label{table:mix}
\begin{tabular}{lllllr}
\hline & & \(JAAD_{all} \) & \(JAAD_{beh} \) &
\(PIE_{all} \) & \(PIE_{beh} \)\\
\cline{3-6}
Conf    & R & F1 & F1 & F1 & F1  \\
\hline
J14K     & 58  & \textbf{0.75}  & \textbf{0.76}  & 0.46  & 0.51 \\
P80K    & 36  & 0.59  & 0.60  & \textbf{0.54}  & 0.57    \\
J8K-P8K   & 61  & 0.73  & 0.74  & 0.52  & 0.57   \\
J10K-P10K & 64  & 0.73  & 0.73  & \textbf{0.54}  & \textbf{0.58}    \\
J14K-P80K & 62  & 0.51  & 0.51  & 0.50  & 0.52   \\
\hline
\end{tabular}
\end{table}

\section{Baseline For Explainability}
\subsection{Results}
In the Table \ref{table:fresults} we gathered the results obtained with our best fuzzy rules generated. In the case of the explainable crossing predictor, the best results over each dataset include the fuzzy rules trained with: 14,000 records from JAAD ordered videos (J14K), 14,000 records from JAAD random videos (J14KR3) and 10,000 records from PIE + 10,000 records from JAAD (J10K-P10K).
\begin{table}[h!]
    \centering
    \caption{Final Results. Abbr. Conf:Configuration, R:Rules, F1: F1 score}
    \label{table:fresults}
\begin{tabular}{lllllr}
\hline &  & \(JAAD_{all} \) & \(JAAD_{beh} \) &
\(PIE_{all} \) & \(PIE_{beh} \)\\
\cline{3-6}
Conf   & Factor  & F1 & F1 & F1 & F1  \\
\hline
J10K-P10K  & Mix & 0.73  & 0.73  & 0.54  & 0.58   \\
J8KR3  & Random & 0.76  & 0.77  & 0.48  & 0.53   \\
J14K  & Quantity & 0.75  & 0.76  & 0.46  & 0.51   \\
\hline
\end{tabular}
\end{table}

\begin{figure}[h!]
  \centering
  \includegraphics[scale=0.28]{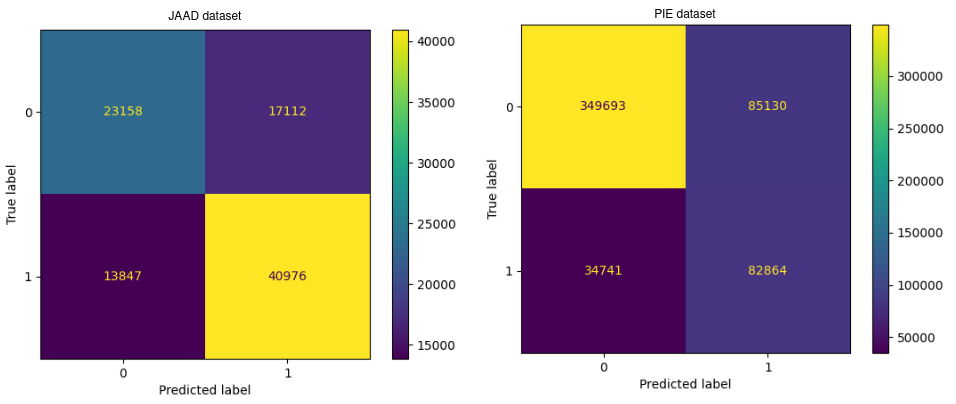}
  \centering
  \caption{Confusion matrix for J10K-P10K configuration}
  \label{fig:confusion matrix}
\end{figure}

It is important to highlight that the results from these configurations are not associated with the condition that the ExPedCross fails in the majority of just one pedestrian behaviour (Crossing or Not Crossing). For instance, the confusion matrix for the configuration J10K-P10K (See Figure \ref{fig:confusion matrix}) evaluated on \(JAAD_{all} \) \(PIE_{all} \) evidences that the mispredictions are balanced between the two classes. That means that the F1 score is not affected because the predictor succeeds only with one of the classification types.

\subsection{Predictions explainability}

Taking as a baseline the best configuration J10K-P10K from JAAD and PIE dataset, we conducted an analysis focused on the rules that ExPedCross activates when predicting pedestrian crossing behaviour within the \(JAAD_{all} \) and \(PIE_{all} \).
Out of the 64 rules that constitute the J10-P10 configuration, all of them were activated during the predictions. However, it was evident that certain rules were triggered more frequently than others. 

In the following figures, we present the top ten most frequently activated rules for predicting pedestrian crossings within the two datasets:

\begin{figure}[h!]
  \centering
  \includegraphics[scale=0.45]{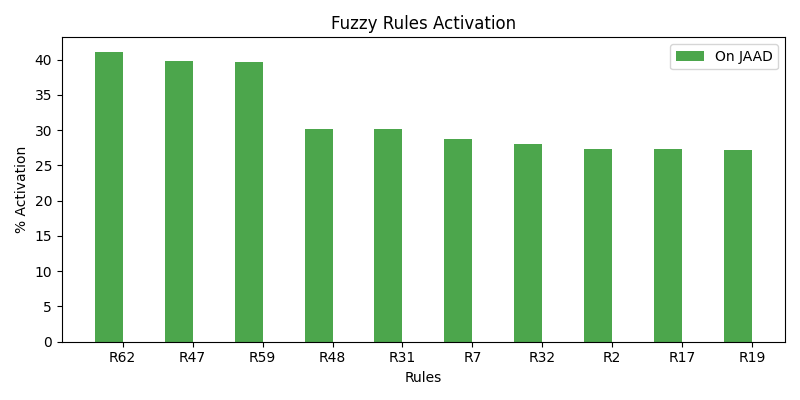}
  \centering
  \caption{Fuzzy rules activation in JAAD dataset}
  \label{fig:jaad_rules_activation}
\end{figure}

In figure \ref{fig:jaad_rules_activation}, it is evident that three primary rules (Rule 62, Rule 47, and Rule 59) were activated the most frequently, while the remaining seven rules exhibited a similar percentage of activation. Regarding the most triggered rule, rule No. 62 was triggered 41.11\% of the time, and its definition is provided above:

 \begin{multline*}
  IF \;( CrossLevel \;IS \;easy ) \;AND (\; Distance \;IS \\
  too \;near)\; AND \;( \; Action \;IS \;run) \;THEN \\
 (Cross \;IS \;crossing) \;WEIGHT\; 0.66
    \end{multline*}

In terms of explainability, the mentioned rule makes sense, considering the three features that can identify a pedestrian who is running and observes that the ego-vehicle is nearby (likely stopped) and in the traffic scene there is a zebra crossing. These conditions collectively suggest a higher probability of crossing. Concerning the other rules within the top 10, it's worth noting that all features are incorporated into these rules. However, the orientation and distance features appear to be more prominently utilized in the rules that were activated.

In contrast, within the \(PIE_{all} \), there exists a notable disparity between the most activated rule and the other nine.

\begin{figure}[h!]
  \centering
  \includegraphics[scale=0.45]{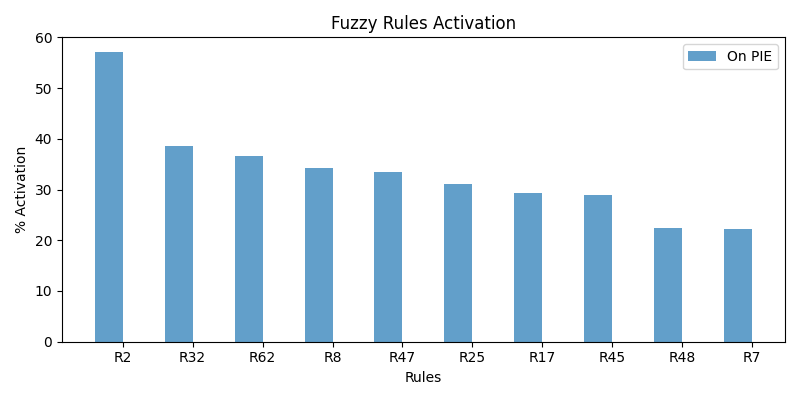}
  \centering
  \caption{Fuzzy rules activation in PIE dataset}
  \label{fig:pie_rules_activation}
\end{figure}

In fact, rule No.2 was the most activated with 57.15\% times:
    \begin{multline*}
        IF \;( Proximity \;IS\; moderate)\; THEN \\
         (Cross\; IS\; not\;crossing)\; with\; WEIGHT\; 0.56
    \end{multline*}

It can be inferred that a moderate proximity of the pedestrian to the road is considered a relevant factor when predicting that a pedestrian will not cross in the PIE dataset. In terms of explainability, individuals who are not close to the road are typically not perceived as pedestrians likely to cross imminently.

Additionally, it is important to highlight that all the extracted features are included in the top 10 rules. Nevertheless, the orientation and distance features appear to be the most frequently used in these rules.

In summary, while the top rules activated for predictions in each dataset differ, seven rules coincide with varying percentages of activation. However, the most influential features for both datasets were consistently the proximity, orientation and distance.

\section{Guidelines and Recommendations}
The following section presents some guidelines and recommendations generated as a result of the experimental insights of the explainable pedestrian crossing predictor. These guidelines include the process of dataset selection and feature selection.

\subsection{Dataset Selection}
During the experiments we identify how important it is to take care and go deeply in the dataset selection process. Therefore we define the following guidelines to choose the adequate dataset for a ML project, especially in the context of autonomous driving and explainability:

\begin{itemize}
    \item \textbf{What you expect from dataset?:} Identify the task you want to achieve and what you expect in detail from the dataset. We recommend to define a checklist of criteria in order to save time when you are selecting the dataset.
    \item \textbf{Detail deeply the dataset:} Take a considerable time to identify, analyse and understand the dataset, its data, properties, videos, images and what it is composed of. Detail how the dataset is labelled and read in detail the documentation.
    \item \textbf{Evaluate quality Vs quantity:} In terms of explainability, a big amount of data for pedestrian crossing task does not guarantee a fuzzy rule generalization on its own. Therefore, it is important to evaluate the quality of the dataset, the diversity of scenes and actors. It can be at the end more relevant than the quantity of data.
    \item \textbf{Each dataset is different:} Take seriously that each dataset is different. Using different datasets with the same preprocessing process and without detailing the differences can lead to unexpected results.
    \item \textbf{Features Preprocessing:} in the way of the explainability it is important to analyse the data incorporated into the dataset and define what will be the data which you will use in your ML process.
    
\end{itemize}

\subsection{Feature Selection}
Regarding the feature selection in the pedestrian crossing prediction task, we got some insights from this work and cluster them to provide some guidelines which can help to get favourable results over this task, especially when the target is to make these systems explainable: 
\begin{itemize}
    \item \textbf{Group the Features:} One feature by itself will not be enough to identify the pedestrian crossing behaviour. It is highly recommended to combine two or more features. The best strategy implies starting with few features and adding the other ones progressively.
    \item \textbf{Identify the generic Features:} There are features which just can be used for a specific dataset, others that need some changes into the preprocessing phase and others that are not dependent from the dataset. So, carefully analyse the use of these features because, it could affect the cross-dataset evaluation.
    \item \textbf{Features Preprocessing:} In order to improve the interpretability and explainability of the pedestrian crossing action prediction, it is important to analyse each feature in detail and identify if the feature needs any processing or if the feature extraction has to consider any variable which can affect or create noise into the system.
\end{itemize}

\section{Conclusions and Future work}
In this work, we have presented a novel, interpretable and explainable approach for pedestrian crossing predictor. This approach is based on a neuro-symbolic model using fuzzy logic. The experiments addressed some evaluation factors which allow to define some guidelines and recommendations regarding the process of data selection and feature selection over the explainable and interpretable context. Through the experiments we emphasize that emphasize that the process of selecting the right dataset, one that is suitable, accurate, and comprehensive, is indeed a challenging task within the domain of ML context.

From these insights, it is important to highlight that in the context of an explainable predictor, having a large amount of data does not necessarily lead to better results and does not guarantee improved generalization on its own. Additionally, as we look for explainable predictions, it is crucial to include a deep analysis of the videos before using them. Another insight shows us that including data selection and filtering strategies is also important with a view to creating meaningful fuzzy rules.

On the other hand, it is important to mention that the features to use need to be selected carefully and one feature by itself could be not enough for getting accurate results; the features need to be complemented with additional features that convey more representative information. Likewise, the divergent representation of these features across datasets may lead to inconsistencies in prediction performance when the explainable predictor is applied, therefore, each feature has to be analyzed and used for each dataset as in an optimal way.

Regarding the use of different features extracted from JAAD and PIE dataset in an explainable approach, the proximity, orientation, action and distance are presented as strong features which can reveal meaningful information about the pedestrian behaviour.   

In the future, we will explore more complex strategies and approaches to develop an explainable and interpretable pedestrian crossing action predictor. In addition, we will work over the features extraction including features which can encapsulate the pedestrian history by the time. 

\section*{Acknowledgments}
This research has been funded by the HEIDI project of the European Commission under Grant Agreement: 101069538

\printbibliography

\end{document}